\documentclass[acmtog]{acm}
\usepackage{color}
\usepackage{subfigure}
\usepackage{units}
\usepackage{expl3}
\usepackage{amsmath}
\usepackage{multirow}
\usepackage{colortbl}

\definecolor{Yellow}{rgb}{1,1, 0.6}
\definecolor{Red}{rgb}{1, 0.6, 0.6}
 
\usepackage{booktabs}
\setcitestyle{square}

\renewcommand\footnotetextcopyrightpermission[1]{}
\makeatletter
\renewcommand\@formatdoi[1]{\ignorespaces}
\makeatother

\begin{document}

\title{High-resolution Deep Convolutional Generative Adversarial Networks}

\author{J. D. Curt\'o$^{*,1,2,3,4}$, I. C. Zarza$^{*,1,2,3,4}$, F. Torre$^{2,5}$, I. King$^{1}$, and M. R. Lyu$^{1}$.}
\affiliation{\institution{\\$^{1}$The Chinese University of Hong Kong. $^{2}$Carnegie Mellon. \\
$^{3}$Eidgen\"ossische Technische Hochschule Z\"urich. 
  $^{4}$City University of Hong Kong. $^{5}$Facebook. \\ $^{*}$Both authors contributed equally.}}

\renewcommand{\shortauthors}{Curt\'o, Zarza, Torre, King, and Lyu.}
\renewcommand{\shorttitle}{High-resolution Deep Convolutional Generative Adversarial Networks.}

\authorsaddresses{\{curto,zarza,king,lyu\}@cse.cuhk.edu.hk, ftorre@cs.cmu.edu \\ \href{https://www.decurto.tw}{decurto.tw} \href{https://www.dezarza.tw}{dezarza.tw}}

\begin{teaserfigure}
\centering
\includegraphics[width=\textwidth]{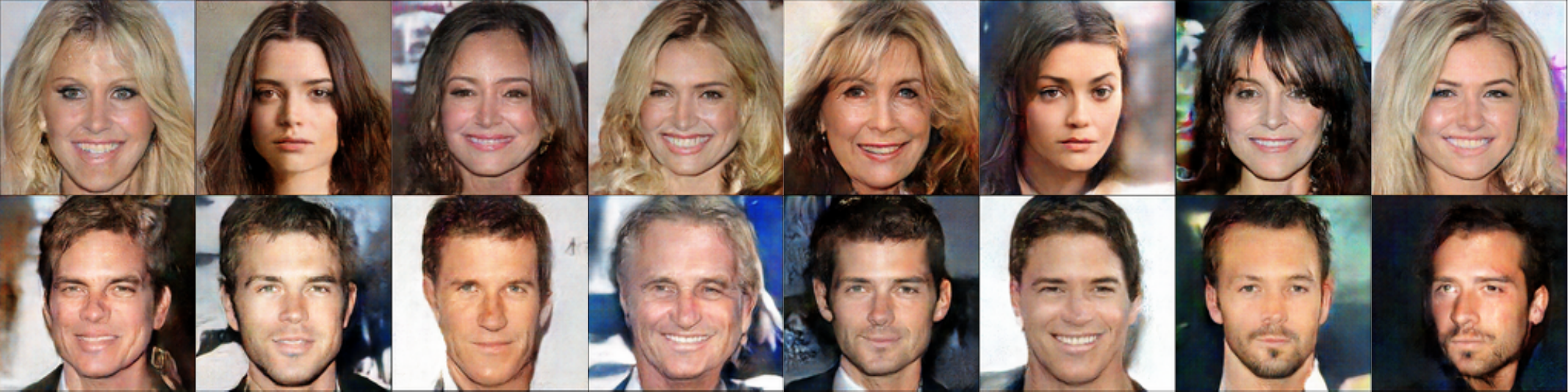}
   \caption{\textbf{HDCGAN Synthetic Images}. A set of random samples. Our system generates high-resolution synthetic faces with an extremely high level of detail. HDCGAN goes from random noise to realistic synthetic pictures that can even fool humans. To demonstrate this effect, we create the Dataset of Curt\'o \& Zarza, the first GAN augmented dataset of faces.}
   \label{fgr:celeba_hq}
\end{teaserfigure}

\ExplSyntaxOn
\newcommand\latinabbrev[1]{
  \peek_meaning:NTF . {
    #1\@}%
  { \peek_catcode:NTF a {
      #1.\@ }%
    {#1.\@}}}
\ExplSyntaxOff

\def\eg{e.g. }
\def\etal{et al. }


\begin{abstract}

Generative Adversarial Networks (GANs) \cite{Goodfellow14} convergence in a high-resolution setting with a computational constrain of GPU memory capacity has been beset with difficulty due to the known lack of convergence rate stability. In order to boost network convergence of DCGAN (Deep Convolutional Generative Adversarial Networks) \cite{Radford16} and achieve good-looking high-resolution results we propose a new layered network, HDCGAN, that incorporates current state-of-the-art techniques for this effect. Glasses, a mechanism to arbitrarily improve the final GAN generated results by enlarging the input size by a telescope $\zeta$ is also presented. A novel bias-free dataset, Curt\'o \& Zarza\footnote{Curt\'o is available at \href{https://www.github.com/curto2/c/}{https://www.github.com/curto2/c/}} \footnote{Code is available at \href{https://www.github.com/curto2/graphics/}{https://www.github.com/curto2/graphics/}}, containing human faces from different ethnical groups in a wide variety of illumination conditions and image resolutions is introduced. Curt\'o is enhanced with HDCGAN synthetic images, thus being the first GAN augmented dataset of faces. We conduct extensive experiments on CelebA \cite{Liu15}, CelebA-hq \cite{Karras18} and Curt\'o. HDCGAN is the current state-of-the-art in synthetic image generation on CelebA achieving a MS-SSIM of 0.1978 and a FR\'ECHET Inception Distance of 8.44.
\end{abstract}


\maketitle
\fancyfoot{}
\thispagestyle{empty}


\section{Introduction}

Developing a Generative Adversarial Network (GAN) \cite{Goodfellow14} able to produce good quality high-resolution samples from images has important applications \cite{Wu16,Yang17,Zhu17,Bousmalis17,Li17,Wang18,Wang18_2,Portenier18,Lombardi18,Yu18,Sankaranarayanan18,Romero18,Chen19} including image inpainting, 3D data, domain translation, video synthesis, image edition, semantic segmentation and semi-supervised learning.
\\

\begin{figure}
\centering
   \includegraphics[scale=0.12]{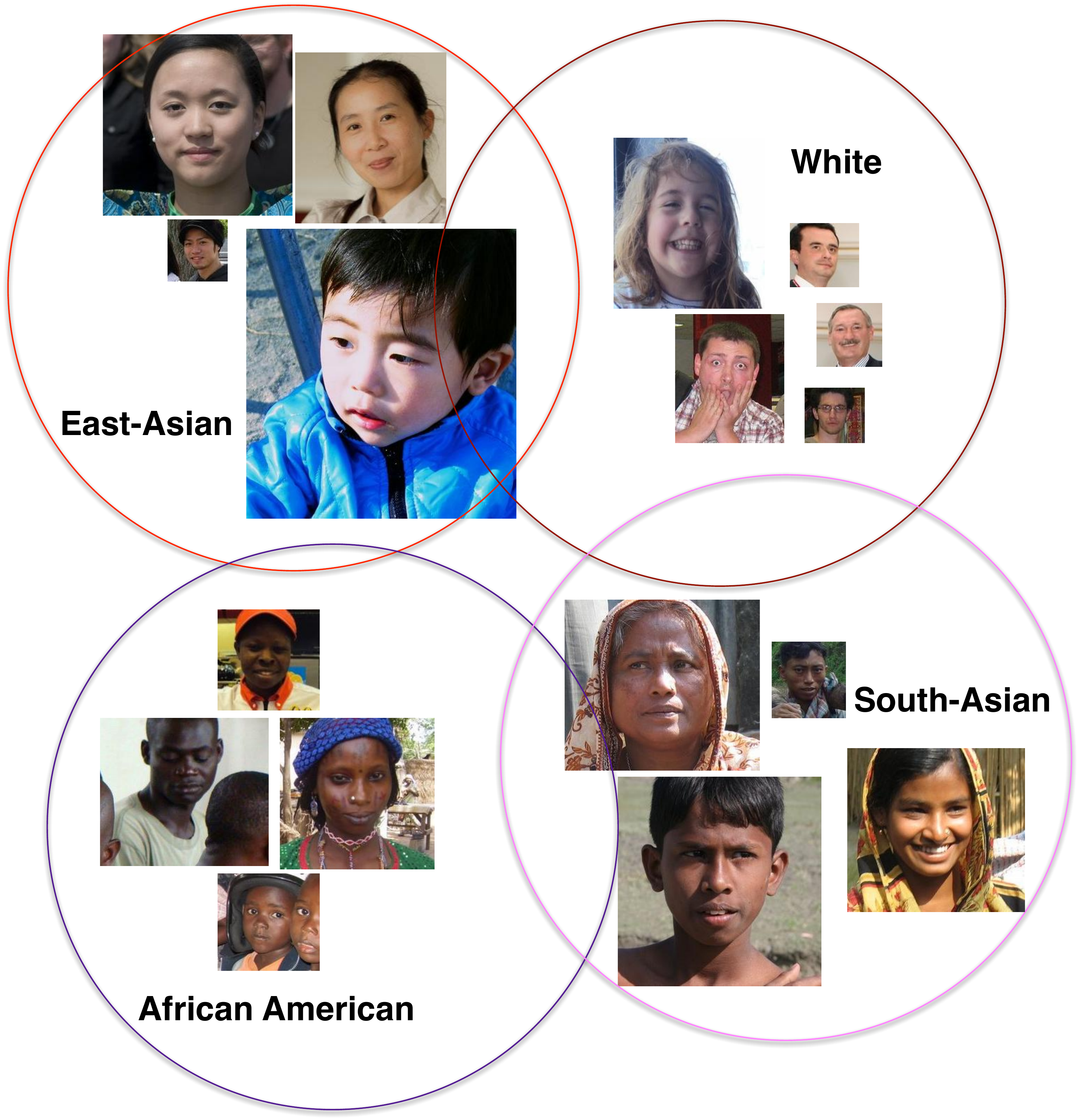}
   \caption{\textbf{Samples of Curt\'o}. A set of random instances for each class of ethnicity: African American, White, East-asian and South-asian. See Table \ref{tbl:curto} for numerics.}
\label{fgr:c}
\end{figure}

In this paper, we focus on the task of face generation, as it gives GANs a huge space of learning attributes. In this context, we introduce the Dataset of Curt\'o \& Zarza, a well-balanced collection of images containing 14,248 human faces from different ethnical groups and rich in a wide range of learnable attributes, such as gender and age diversity, hair-style and pose variation or presence of smile, glasses, hats and fashion items. We also ensure the presence of changes in illumination and image resolution. We propose to use Curt\'o as de facto approach to empirically test the distribution learned by a GAN, as it offers a challenging problem to solve, while keeping the number of samples, and therefore training time, bounded. It can also be used as a drop-in substitute of MNIST for simple tasks of classification, say for instance using labels of ethnicity, gender, age, hair style or smile. It ships with scripts in TensorFlow and Python that allow benchmarks of classification. A set of random samples can be seen in Figure \ref{fgr:c}.\\

Despite improvements in GANs training stability \cite{Salimans16,Mescheder17,Mescheder18} and specific-task design during the last years, it is still challenging to train GANs to generate high-resolution images due to the disjunction in the high dimensional pixel space between supports of the real image and implied model distributions \cite{Arjovsky17,Sonderby17}.
\\

Our goal is to be able to generate indistinguishable sample instances using face data to push the boundaries of GAN image generation that scale well to high-resolution images (such as 512$\times$512) and where context information is maintained. 
\\ 

In this sense, Deep Learning has a tremendous appetite for data. The question that arises instantly is, what if we were able to generate additional realistic data to aid learning using the same techniques that are later used to train the system. The first step would then be to have an image generation tool able to sample from a very precise distribution (\eg faces from celebrities) which instances resemble or highly correlate with real sample images of the underlying true distribution. Once achieved, what is desirable and comes next is that these generated image points not only fit well into the original distribution set of images but also add additional useful information such as redundancy, different poses or even generate highly-probable scenarios that would be possible to see in the original dataset but are actually not present. 
\\

Current research trends link Deep Learning and Kernel Methods to establish a unifying theory of learning \cite{Curto17}. The next frontier in GANs would be to achieve learning at scale with very few examples. 
To achieve the former goal this work contributes in the following:
\begin{itemize}
\item Network that achieves compelling results and scales well to the high-resolution setting where to the best of our knowledge the majority of other variants are unable to continue learning or fall into mode collapse.\\  
\item New dataset targeted for GAN training, Curt\'o, that introduces a wide space of learning attributes. It aims to provide a well-posed difficult task while keeping training time and resources tightly bounded to spearhead research in the area.
\end{itemize}


\section{Prior Work}

Generative image generation is a key problem in Computer Vision and Computer Graphics. Remarkable advances have been made with the renaissance of Deep Learning. Variational Autoencoders (VAE) \cite{Kingma14,Lombardi18} formulate the problem with an approach that builds on probabilistic graphical models, where the lower bound of data likelihood is maximized. Autoregressive models (scilicet PixelRNN \cite{Oord16}), based on modeling the conditional distribution of the pixel space, have also presented relative success generating synthetic images. Lately, Generative Adversarial Networks (GANs) \cite{Goodfellow14,Radford16,Odena17,Antoniou18,Wang16,Zhu16,Portenier18} have shown strong performance in image generation. However, training instability makes it very hard to scale to high-resolution (256$\times$256 or 512$\times$512) samples. Some current works on the topic pinpoint this specific problem \cite{Zhang17}, where conditional image generation is also tackled while other recent techniques \cite{Salimans16,Chen17,Dosovitskiy16,Zhao17,Karras18,Wei18,Brock19} try to stabilize training. 


\section{Dataset of Curt\'o \& Zarza}

Curt\'o contains 14,248 faces balanced in terms of ethnicity: African American, East-Asian, South-Asian and White. Mirror images are included to enhance pose variation and there is roughly $25\%$ per image class. Attribute information, see Table \ref{tbl:curto}, is composed of thorough labels of gender, age, ethnicity, hair color, hair style, eyes color, facial hair, glasses, visible forehead, hair covered and smile.
There is also an extra set with 3,384 cropped labeled images of faces, ethnicity white, no mirror samples included, see Column 4 in Table \ref{tbl:curto} for statistics. We crawled Flickr to download images of faces from several countries that contain different hair-style variations and style attributes. These images were then processed to extract 49 facial landmark points using \cite{Xiong13}. We ensure using Mechanical Turk that the detected faces are correct in terms of ethnicity and face detection. Cropped faces are then extracted to generate multiple resolution sources. Mirror augmentation is performed to further enhance pose variation.
\\

Curt\'o introduces a difficult paradigm of learning, where different ethnical groups are present, with very varied fashion and hair styles. The fact that the photos are taken using non-professional cameras in a non-controlled environment, gives us multiple poses, illumination conditions and camera quality.

\setlength{\tabcolsep}{4pt}
\begin{table}
\begin{center}
\caption{Dataset of Curt\'o \& Zarza. Attribute Information. Descending order of class instances by number of samples, Column 3. }
\label{tbl:curto}
\begin{tabular}{llll}
\hline\noalign{\smallskip}
  Attribute & Class & \# Samples & \# Extra\\
\noalign{\smallskip}
\hline
\noalign{\smallskip}
Age & Early Adulthood & 3606 & 966\\
& Middle Aged & 2954 & 875\\
& Teenager & 2202 & 178\\
& Adult & 1806 & 565\\
& Kid & 1706 & 85 \\
& Senior & 1102 & 402\\
& Retirement & 436 & 218\\
& Baby & 232 & 14\\
\hline
Ethnicity & African American & 4348 & 0\\
& White & 3442 & 3384\\
& East Asian & 3244 & 0\\
& South Asian & 3214 & 0\\
\hline
Eyes Color & Brown & 9116 & 2119\\
& Other & 4136 & 875\\
& Blue & 580 & 262\\
& Green & 416 & 128\\
\hline
Facial Hair & No & 12592 & 2821\\
& Light Mustache & 466 & 156\\
& Light Goatee & 444 & 96\\
& Light Beard & 258 & 142\\
& Thick Goatee & 168 & 39\\
& Thick Beard & 166 & 68\\
& Thick Mustache & 154 & 62\\
\hline
Gender & Male & 7554 & 1998\\
& Female & 6694 & 1386\\
\hline
Glasses & No & 12576 & 2756\\
& Eyeglasses & 1464 & 539\\
& Sunglasses & 208 & 89\\
\hline
Hair Color & Black & 8402 & 964\\
& Brown & 3038 & 1241\\
& Other & 1554 & 253\\
& Blonde & 616 & 543\\
& White & 590 & 347\\
& Red & 48 & 36\\
\hline
Hair Covered & No & 12292 & 3060\\
& Turban & 1206 & 76\\
& Cap & 722 & 237\\
& Helmet & 28 & 11\\
\hline
Hair Style & Short Straight & 5038 & 1642\\
& Long Straight & 2858 & 857\\
& Short Curly & 2524 & 287\\
& Other & 2016 & 249\\
& Bald & 1298 & 187\\
& Long Curly & 514 & 162\\
\hline
Smile & Yes & 8428 & 2118\\
& No & 5820 & 1266\\
\hline
Visible Forehead & Yes & 11890 & 3033\\
& No & 2358 & 351\\
\hline
\end{tabular}
\end{center}
\end{table}
\setlength{\tabcolsep}{1.4pt}


\section{Approach}

\begin{figure}[H]
\centering
   \includegraphics[scale=0.2]{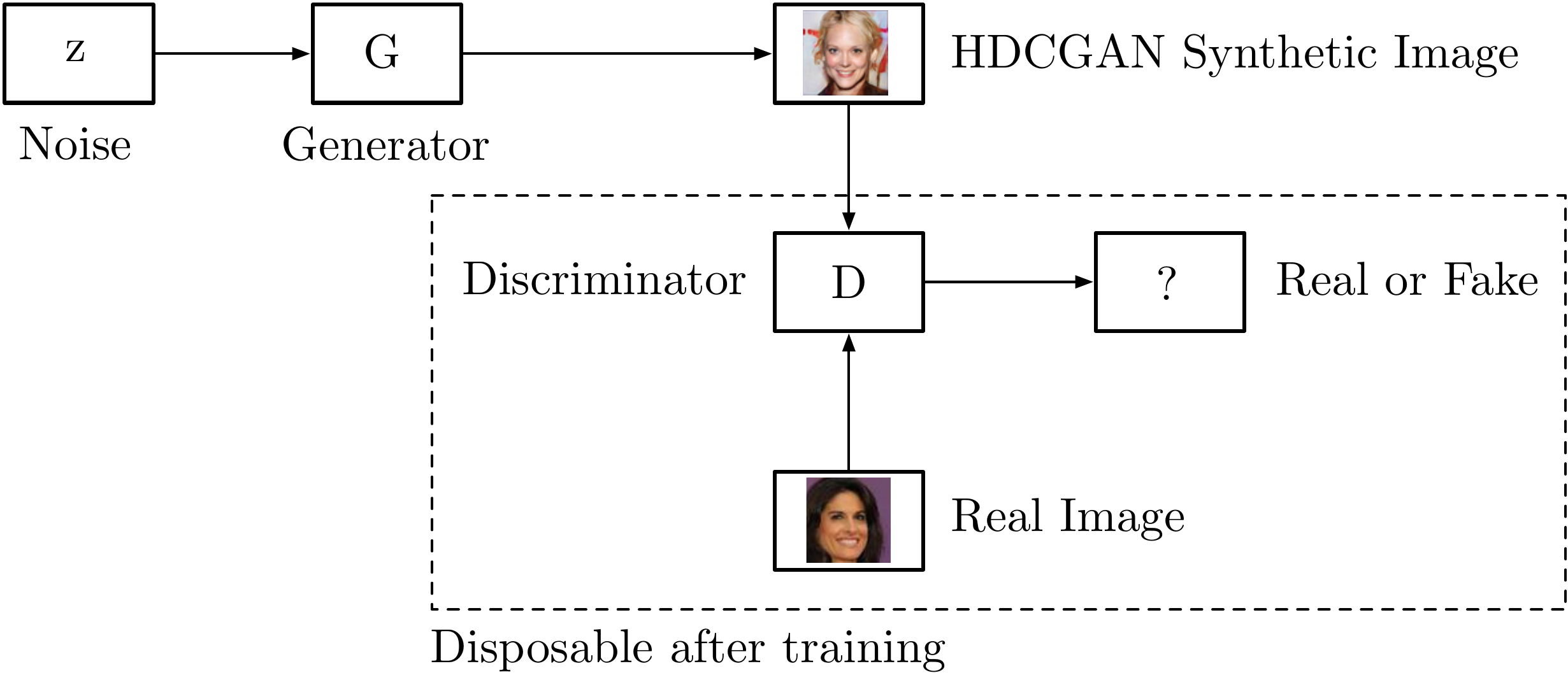}
   \caption{\textbf{Generative Adversarial Networks}. A two-player game between the Generator $G$ and the Discriminator $D$. The dotted line denotes elements that will not be further used after the game stops, namely, end of training.}
\label{fgr:gan}
\end{figure}

Generative Adversarial Networks (GANs) proposed by \cite{Goodfellow14} are based on two dueling networks, Figure \ref{fgr:gan}; Generator $G$ and Discriminator $D$. In essence, the process of learning consists of a two-player game where $D$ tries to distinguish between the prediction of $G$ and the ground truth, while at the same time $G$ tries to fool $D$ by producing fake instance samples as closer to the real ones as possible. The solution to a game is called NASH equilibrium.

The min-max game entails the following objective function

\begin{align}
\min_{G} \max_{D} V(D,G) = \mathbb{E}_{x \sim p_{data}} \left[ \log D(x) \right] +\\
+ \mathbb{E}_{z \sim p_{z}} \left[ \log(1 - D(G(z)))\right],
\end{align}

\noindent
where $x$ is a ground truth image sampled from the true distribution $p_{data}$, and $z$ is a noise vector sampled from $p_{z}$ (that is, uniform or normal distribution).  $G$ and $D$ are parametric functions where $G: p_{z} \rightarrow p_{data}$ maps samples from noise distribution $p_{z}$ to data distribution $p_{data}$. 
\\

The goal of the Discriminator is to minimize

\begin{align}
L^{(D)}&= -\frac{1}{2}\mathbb{E}_{x \sim p_{data}} \left[ \log D(x) \right]-\\
&- \frac{1}{2}\mathbb{E}_{z \sim p_{z}} \left[ \log(1 - D(G(z)))\right].
\end{align}
\\

If we differentiate it w.r.t $D(x)$ and set the derivative equal to zero, we can obtain the optimal strategy

\begin{align}
D(x) = \frac{p_{data}(x)}{p_{z}(x)+p_{data}(x)}.
\end{align}
\\

Which can be understood intuitively as follows. Accept an input, evaluate its probability under the distribution of the data, $p_{data}$, and then evaluate its probability under the generator's distribution of the data, $p_{z}$. Under the condition in $D$ of enough capacity, it can achieve its optimum. Note the discriminator does not have access to the distribution of the data but it is learned through training. The same applies for the generator's distribution of the data. Under the condition in $G$ of enough capacity, then it will set $p_{z} = p_{data}$. This results in $D(x)= \frac{1}{2}$, that is actually the NASH equilibrium. In this situation, the generator is a perfect generative model, sampling from $p(x)$.
\\

As an extension to this framework, DCGAN \cite{Radford16} proposes an architectural topology based on Convolutional Neural Networks (CNNs) to stabilize training and re-use state-of-the-art networks from tasks of classification. This direction has recently received lots of attention due to its compelling results in supervised and unsupervised learning. We build on this to propose a novel DCGAN architecture to address the problem of high-resolution image generation. We name this approach HDCGAN.

\subsection{HDCGAN}

Despite the undoubtable success, GANs are still arduous to train, particularly when we use big images (\eg 512$\times$512). It is very common to see $D$ beating $G$ in the process of learning, or the reverse, ending in unrecognizable imagery, also known as mode collapse. Only when stable learning is achieved, the GAN structure is able to succeed in getting better and better results with time. 
\\

This issue is what drives us to carefully derive a simple yet powerful structure that leverages common problems and gets a stable and steady training mechanism. 
\\

Self-normalizing Neural Networks (SNNs) were introduced in \cite{Klarbauer17}. We consider a neural network with activation function $f$, connected to the next layer by a weight matrix $\mathbf{W}$, and whose inputs are the activations from the preceding layer $x$, $y=f(\mathbf{W}x)$.
\\

We can define a mapping $g$ that maps mean and variance from one layer to mean and variance of the following layer
\begin{align} 
\binom{\mu}{\nu}  \longmapsto \binom{\tilde{\mu}}{\tilde{\nu}} : \binom{\tilde{\mu}}{\tilde{\nu}}  = g \binom{\mu}{\nu}.
\end{align}

Common normalization tactics such as batch normalization ensure a mapping $g$ that keeps $(\mu,\nu)$ and $(\tilde{\mu},\tilde{\nu})$ close to a desired value, normally (0, 1).
\\

SNNs go beyond this assumption and require the existence of a mapping $g:\Omega \longmapsto \Omega$ that for each activation $y$ maps mean and variance from one layer to the next layer and at the same time have a stable and attracting fixed point depending on $(\omega, \tau)$ in $\Omega$. Moreover, the mean and variance remain in the domain $\Omega$ and when iteratively applying the mapping $g$, each point within $\Omega$ converges to this fixed point. Therefore, SNNs keep activations normalized when propagating them through the layers of the network.
\\

Here $(\omega, \tau)$ are defined as follows. For $n$ units with activation $x_{c}$, $1 \leq c \leq n$ in the lower layer, we set $n$ times the mean of the weight vector $w \in \mathbb{R}^{n}$ as $\omega : =\sum_{c=1}^{n}w_{c}$ and $n$ times the second moment as $\tau:=\sum_{c=1}^{n}w_{c}^{2}$.
\\
  
Scaled Exponential Linear Units (SELU) \cite{Klarbauer17} is introduced as the choice of activation function in Feed-forward Neural Networks (FNNs) to construct a mapping $g$ with properties that lead to SNNs.
\begin{align} 
selu(x) = \lambda \left\{ \begin{array}{ll}
 x & \textrm{if $x>0$}\\
 \alpha \exp^{x} - \alpha & \textrm{if $x\leq0$.}
  \end{array} \right.
\end{align}

Empirical observation leads us to say that the use of SELU greatly improves the convergence speed on the DCGAN structure, however, after some iterations mode collapse and gradient explosion completely destroy training when using high-resolution images. We conclude that although SELU gives theoretical guarantees as the optimal activation function in FNNs, numerical errors in the GPU computation degrade its performance in the overall min-max game of DCGAN. To alleviate this problem, we propose to use SELU and BatchNorm \cite{Ioffe15} together. The motivation is that when numerical errors move $(\tilde{\mu},\tilde{\nu})$ away from the attracting point that depends on $(\omega, \tau) \in \Omega$, BatchNorm will ensure it is close to a desired value and therefore maintain the convergence rate. 
\\

Experiments show that this technique stabilizes training and allows us to use fewer GPU resources, having steady diminishing errors in $G$ and $D$. It also accelerates convergence speed by a great factor, as can be seen after some few epochs of training on CelebA in Figure \ref{fgr:hdcgan19}. 

\begin{figure}[H]
\centering
   \includegraphics[scale=0.41]{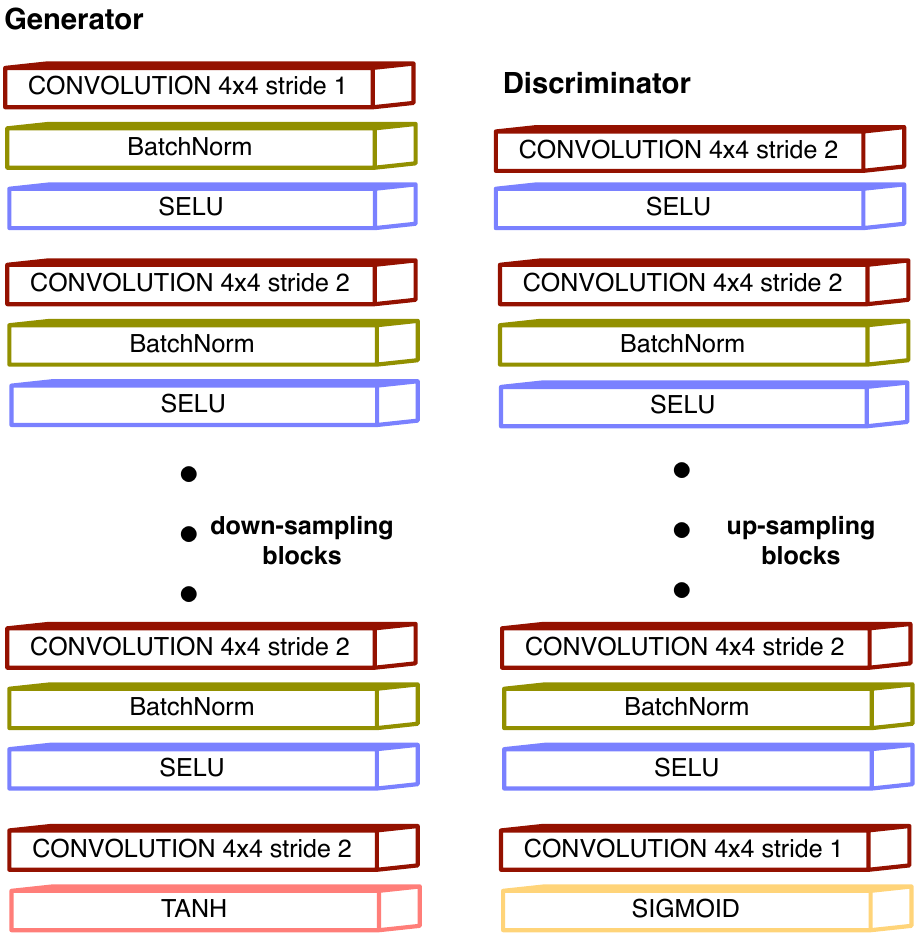}
   \caption{\textbf{HDCGAN Architecture}. Generator and Discriminator.}
\label{fgr:gd}
\end{figure}

As SELU $+$ BatchNorm (BS) layers keep mean and variance close to $(0, 1)$ we get an unbiased estimator of $p_{data}$ with contractive finite variance. These are very desirable properties from the point of view of an estimator as we are iteratively looking for a MVU (Minimum Variance Unbiased) criterion and thus solving MSE (Minimum Square Error) among unbiased estimators. Hence, if the MVU estimator exists and the network has enough capacity to actually find the solution, given a sufficiently large sample size by the Central Limit Theorem, we can attain NASH equilibrium. 
\\

HDCGAN Architecture is described in Figure \ref{fgr:gd}. It differs from traditional DCGAN in the use of BS layers instead of ReLUs. 
\\

We observe that when having difficulty in training DCGAN, it is always better to use a fixed learning rate and instead increase the batch size. This is because having more diversity in training, gives a steady diminishing loss and better generalization. To aid learning, noise following a Normal $N(0,1)$ is added to both the inputs of $D$ and $G$. We see that this helps overcome mode saturation and collapse whereas it does not change the distribution of the original data.
\\

We empirically show that the use of BS induces SNNs properties in the GAN structure, and thus makes learning highly robust, even in the stark presence of noise and perturbations. This behavior can be observed when the zero-sum game problem stabilizes and errors in $D$ and $G$ jointly diminish, Figure \ref{fgr:bs}. Comparison to traditional DCGAN, Wasserstein GAN \cite{Arjovsky17_2} and WGAN-GP \cite{Gulrajani17} is not possible, as to date, the majority of former methods, such as \cite{Denton15}, cannot generate recognizable results in image size 512$\times$512, 24GB GPU memory setting. 
\\

Thus, HDCGAN pushes up state-of-the-art results beating all former DCGAN-based architectures and shows that, under the right circumstances, BS can solve the min-max game efficiently.

\subsection{Glasses}

We introduce here a key technique behind the success of HDCGAN. Once we have a good convergence mechanism for large input samples, that is a concatenation of BS layers, we observe that we can arbitrarily improve the final results of the GAN structure by the use of a Magnifying Glass approach. Assuming our input length is $N \times M$, we can enlarge it by a constant factor, $\zeta_{1} N \times \zeta_{2} M$, which we call telescope, and then feed it into the network, maintaining the size of the convolutional filters untouched. This simple procedure works similar to how contact lenses correct or assist defective eyesight on humans and empowers the GAN structure to appreciate the inner properties of samples.
\\

Note that as the input gets bigger so does the neural network. That is, the number of layers is implicitly set by the image size, see up-sampling and down-sampling blocks in Figure \ref{fgr:gd}. For example, for an input size of $32$ we have $4$ layers while for an input size of $256$ we have $7$ layers. 

\begin{figure}[H]
\centering
\includegraphics[scale=0.14]{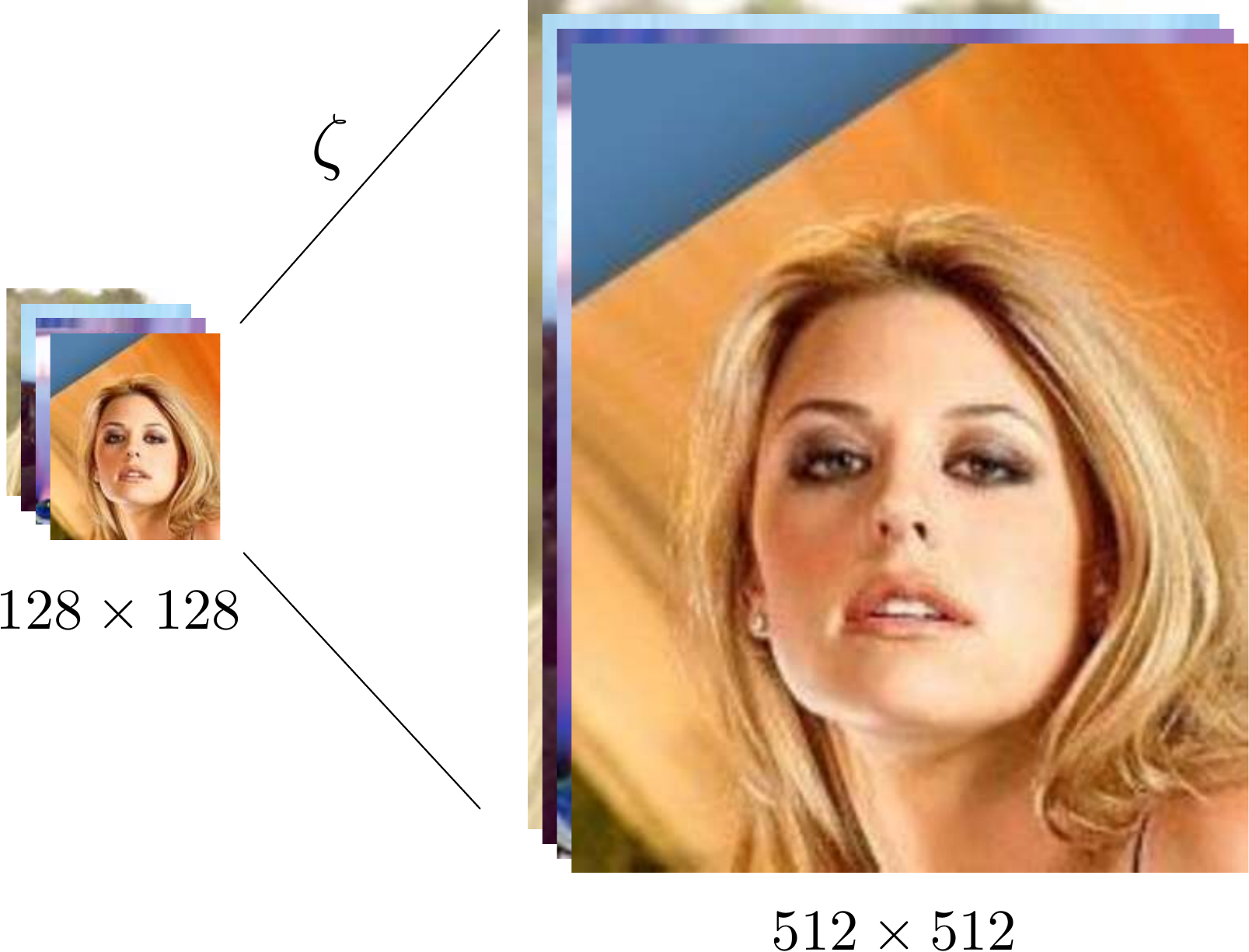}
   \caption{\textbf{Glasses on a set of samples from CelebA}. HDCGAN introduces the use of a Magnifying Glass approach, enlarging the input size by a telescope $\zeta$.}
\label{fgr:hdcgan2}
\end{figure}

We can empirically observe that BS layers together with Glasses induce high capacity into the GAN structure so that a NASH equilibrium can be reached. That is to say, the generator draws samples from $p_{data}$, which is the distribution of the data, and the discriminator is not able to distinguish between them, $D(x)=\frac{1}{2} \forall x$.


\section{Empirical Analysis}

We build on DCGAN and extend the framework to train with high-resolution images using Pytorch. Our experiments are conducted using a fixed learning rate of 0.0002 and ADAM solver \cite{Kingma15} with batch size 32 and 512$\times$512 samples with the number of filters of $G$ and $D$ equal to 64. 
\\

In order to test generalization capability, we train HDCGAN in the newly introduced Curt\'o, CelebA and CelebA-hq. 
\\

Technical Specifications: 2 $\times$ NVIDIA Titan X, Intel Core i7-5820k@3.30GHz. 

\subsection{Curt\'o}

\begin{figure}[H]
\centering
   \includegraphics[scale=0.37]{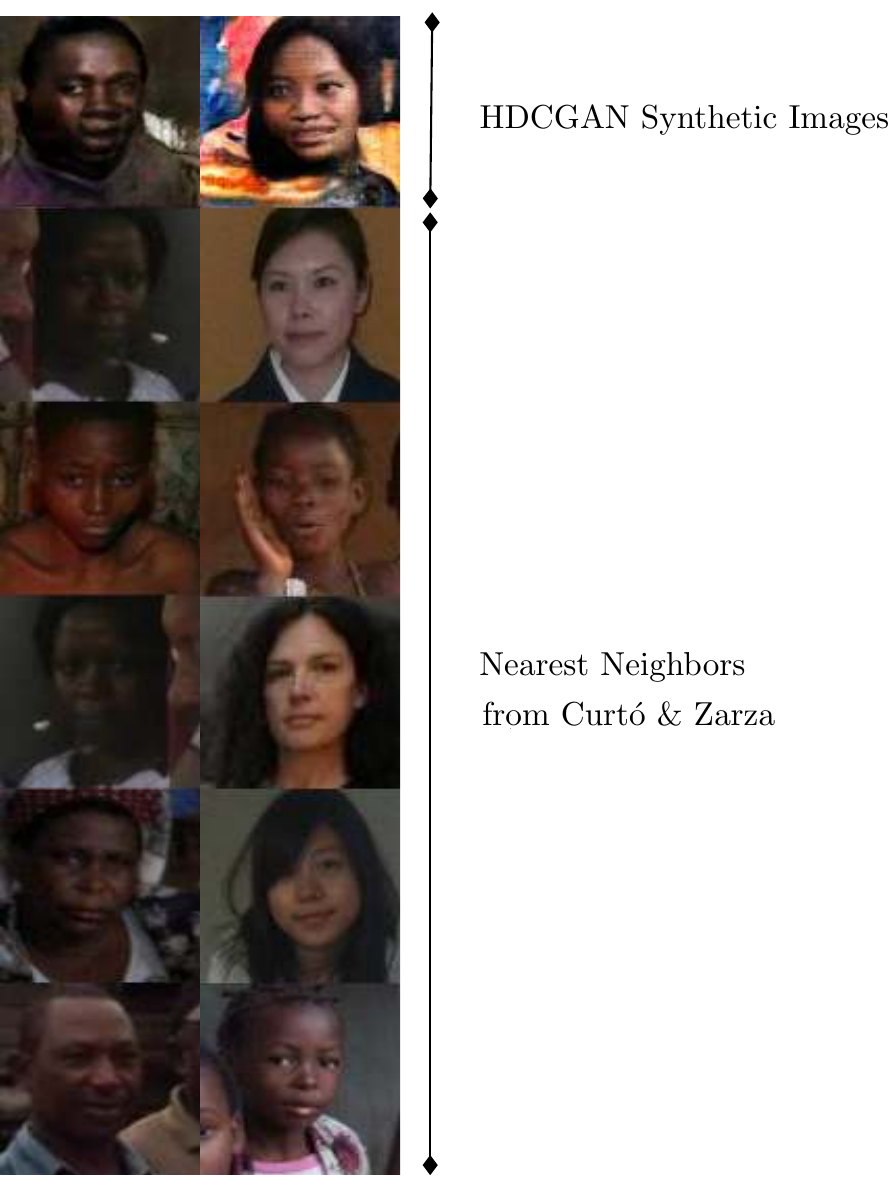}
   \caption{\textbf{Nearest Neighbors. Dataset of Curt\'o \& Zarza}. Generated samples in the first row and their five nearest neighbors in training (rows 2-6).}
\label{fgr:nn}
\end{figure}

The results after 150 epochs are shown in Figure \ref{fgr:hdcgan2}. We can see that HDCGAN captures the underlying features that represent faces and not only memorizes training examples. We retrieve nearest neighbors to the generated images in Figure \ref{fgr:nn} to illustrate this effect.

\begin{figure}[H]
\centering
\includegraphics[scale=0.32]{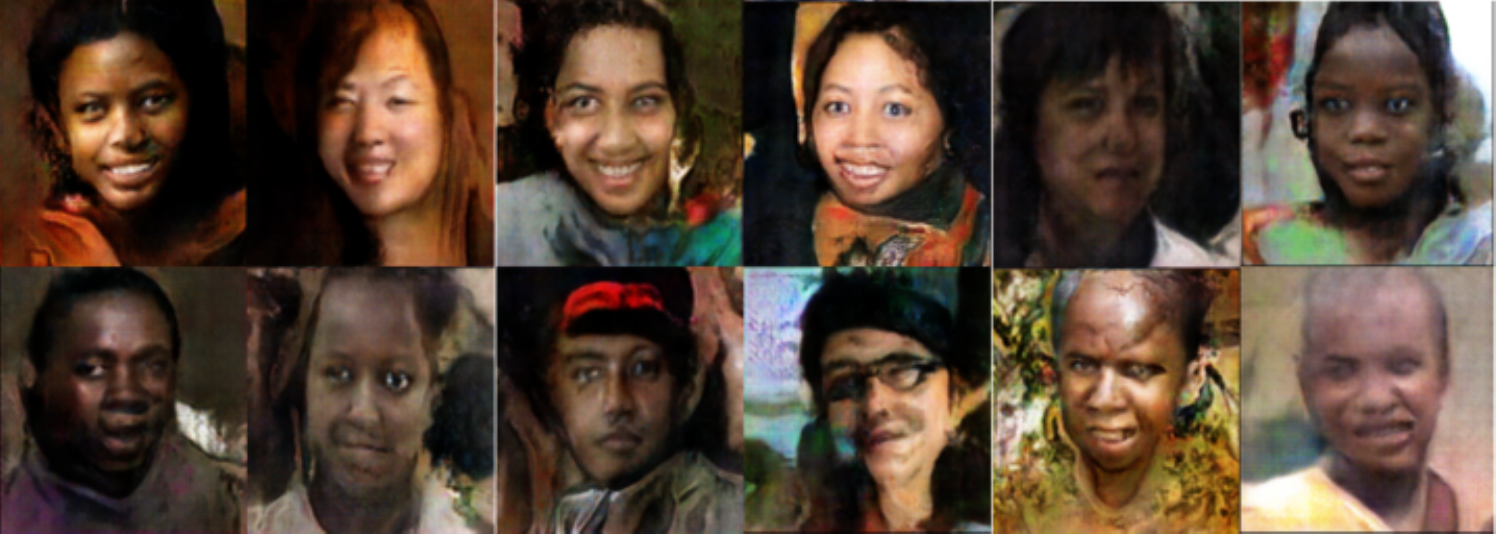}
   \caption{\textbf{HDCGAN Example Results. Dataset of Curt\'o \& Zarza}. 150 epochs of training. Image size 512$\times$512.}
\label{fgr:hdcgan2}
\end{figure}

\subsection{CelebA}

\begin{figure}[H]
\centering
   \includegraphics[scale=0.23]{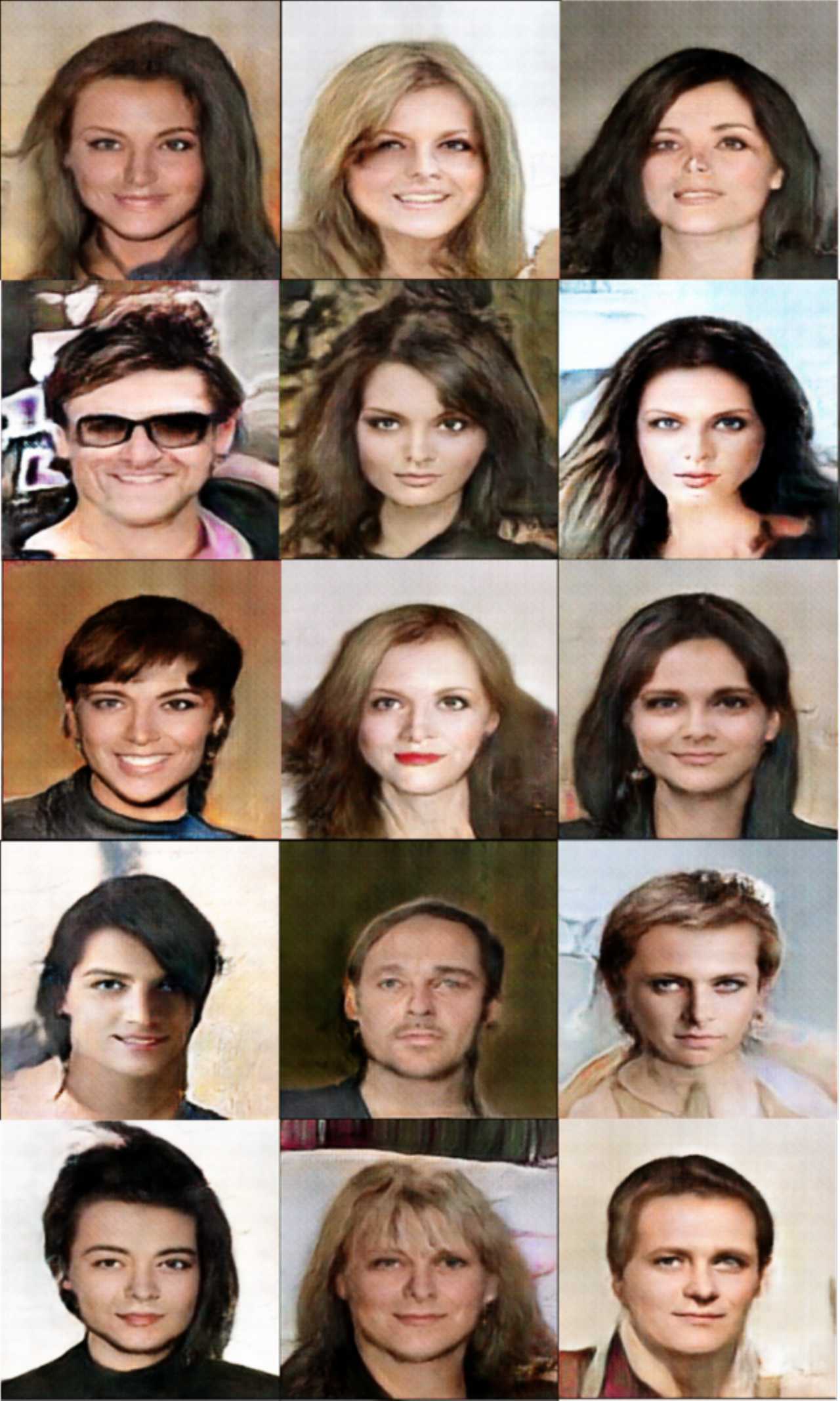}
   \caption{\textbf{HDCGAN Example Results. CelebA}. 19 epochs of training. Image size 512$\times$512. The network learns swiftly a clear pattern of the face.}
\label{fgr:hdcgan19}
\end{figure}

CelebA is a large-scale dataset with 202,599 celebrity faces. It mainly contains frontal portraits and is particularly biased towards groups of ethnicity white. The fact that it presents very controlled illumination settings and good photo resolution, makes it a considerably easier problem than Curt\'o. The results after 19 epochs of training are shown in Figure \ref{fgr:hdcgan19}.

\begin{figure}[H]
\centering
   \includegraphics[scale=0.8]{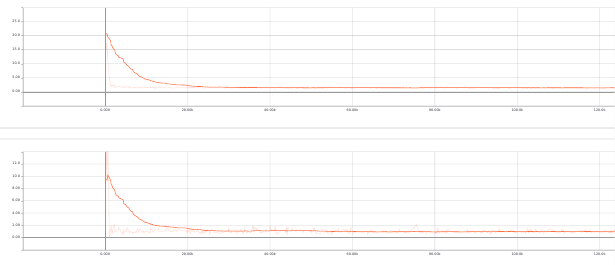}
   \caption{\textbf{HDCGAN on CelebA}. Error in Discriminator (top) and Error in Generator (bottom). 19 epochs of training.}
\label{fgr:bs}
\end{figure}

\begin{figure}[H]
\centering
\includegraphics[scale=0.16]{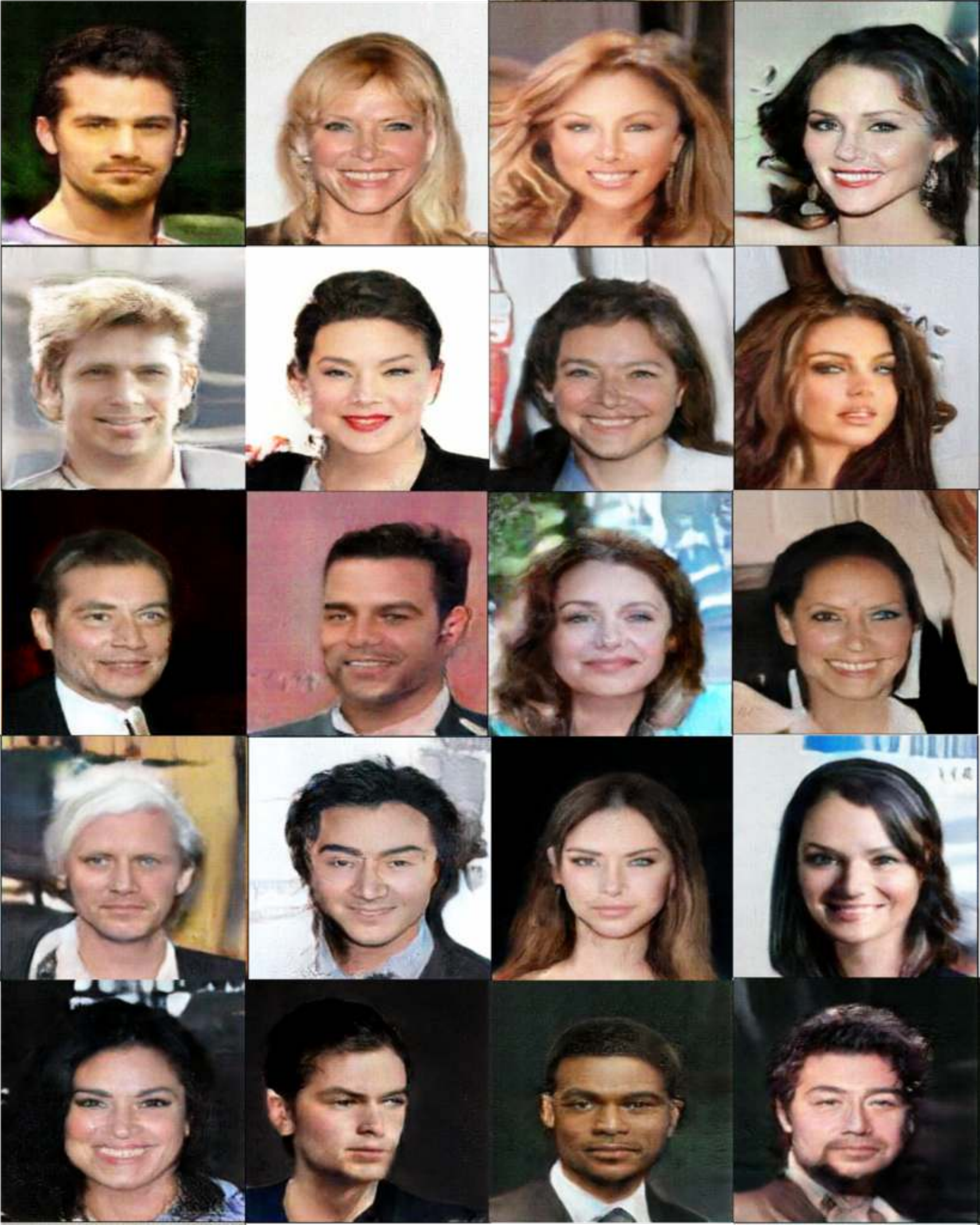}
   \caption{\textbf{HDCGAN Example Results. CelebA}. 39 epochs of training. Image size 512$\times$512. The network generates distinctly accurate and assorted faces, including exhaustive details.}
\label{fgr:hdcgan39}
\end{figure}

\begin{figure}[H]
\centering
\includegraphics[scale=0.27]{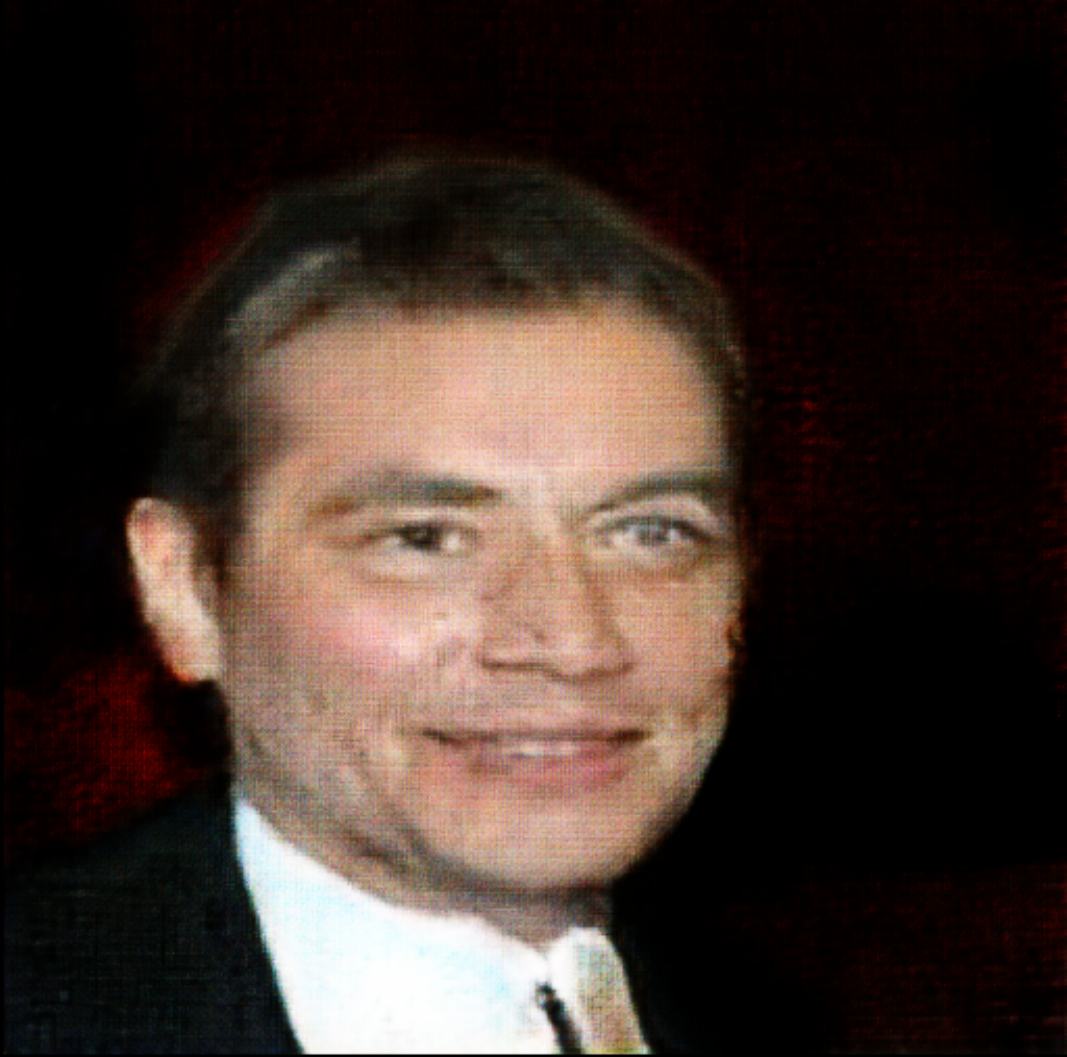}
   \caption{\textbf{HDCGAN Example Result. CelebA}. 39 epochs of training. Image size 512$\times$512. 27\% of full-scale image.}
\label{fgr:hdcgan39_2}
\end{figure}

\begin{figure}[H]
\centering
\includegraphics[scale=0.30]{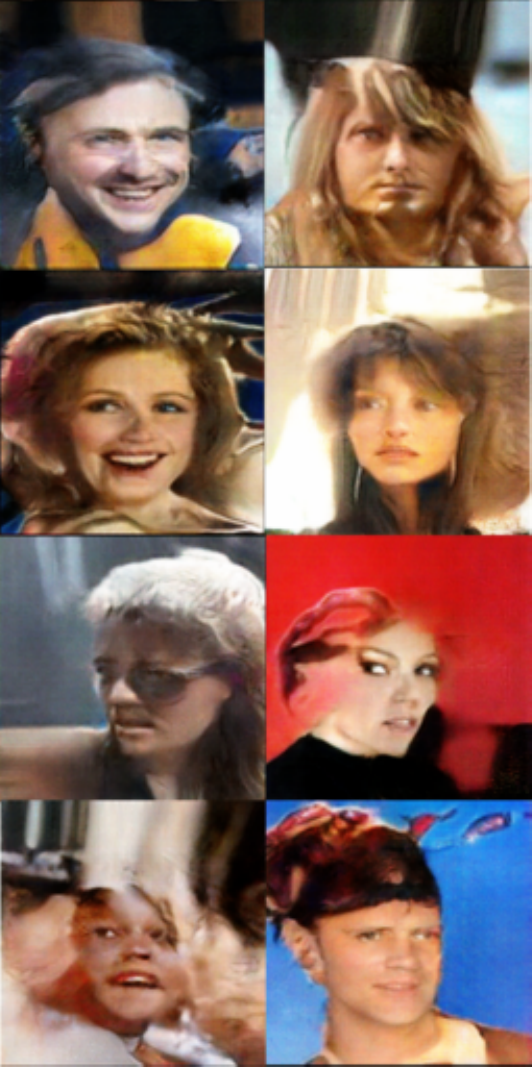}
   \caption{\textbf{HDCGAN Example Results. CelebA}. 39 epochs of training. Image size 512$\times$512. Failure cases. The number of failure cases declines over time, and when present, they are of more meticulous nature.}
\label{fgr:hdcgan39_3}
\end{figure}

In Figure \ref{fgr:bs} we can observe that BS stabilizes the zero-sum game, where errors in $D$ and $G$ concomitantly diminish. To show the validity of our method, we enclose Figure \ref{fgr:hdcgan39}, presenting a large number of samples for epoch 39. We also attach a zoomed-in example to appreciate the quality and size of the generated samples, Figure \ref{fgr:hdcgan39_2}. Failure cases can be observed in Figure \ref{fgr:hdcgan39_3}.

\begin{figure}[H]
\centering
\includegraphics[scale=0.19]{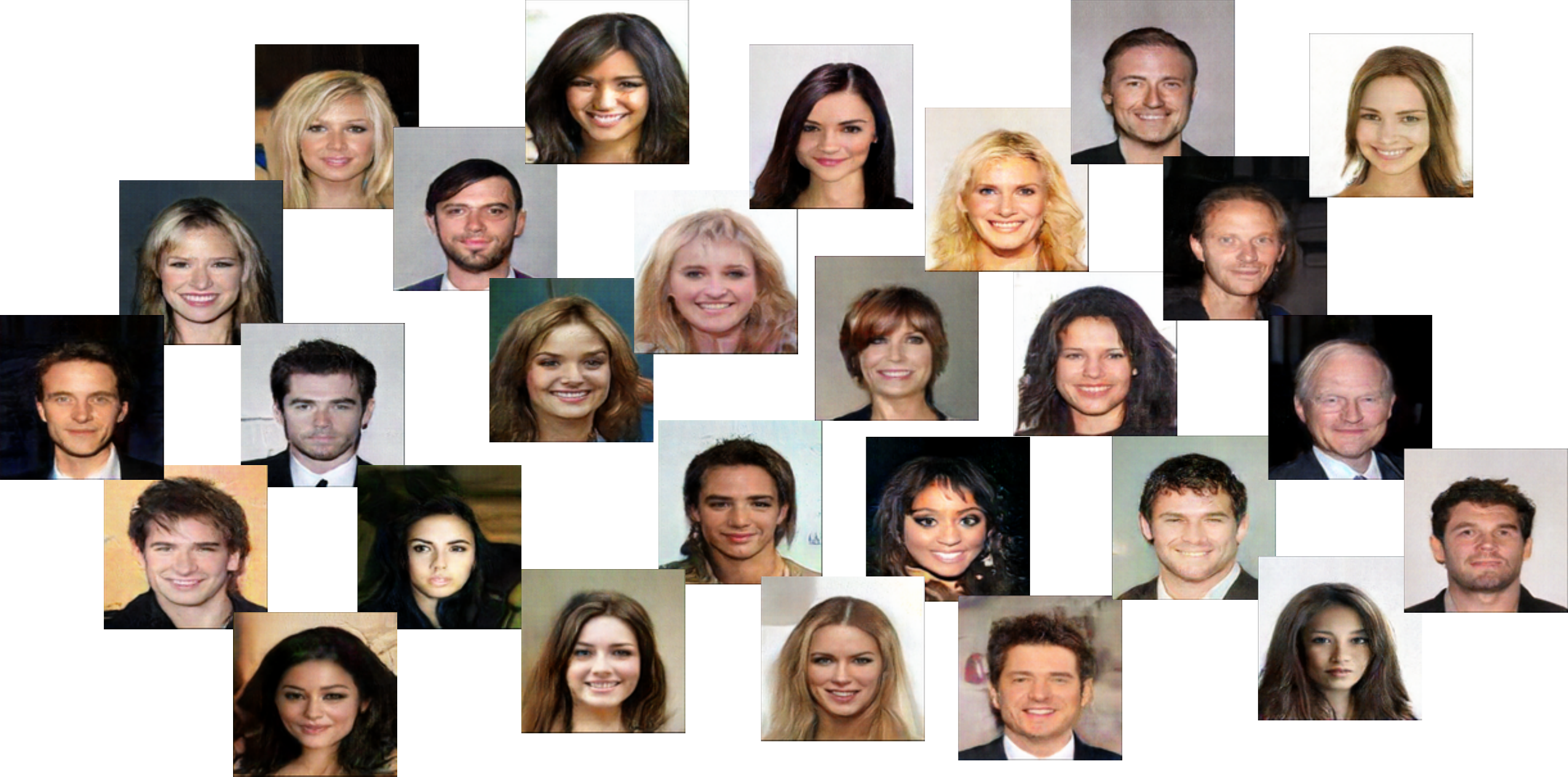}
\caption{\textbf{HDCGAN Synthetic Images}. A set of random samples.}
\label{fgr:hdcgan}
\end{figure}

Besides, to illustrate how fundamental our approach is, we enlarge Curt\'o with 4,239 unlabeled synthetic images generated by HDCGAN on CelebA, a random set can be seen in Figure \ref{fgr:hdcgan}.

\subsection{CelebA-hq}

\begin{figure}[H]
\centering
\includegraphics[scale=0.4]{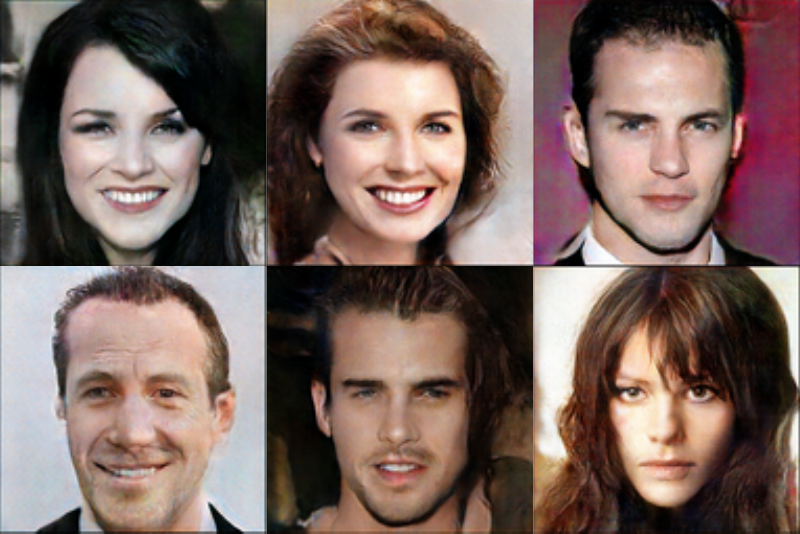}
\caption{\textbf{HDCGAN Example Results. CelebA-hq}. 229 epochs of training. Image size 512$\times$512. The network generates superior faces, with great attention to detail and quality.}
\label{fgr:celeba_hq2}
\end{figure}

\begin{figure}[H]
\centering
\includegraphics[scale=0.27]{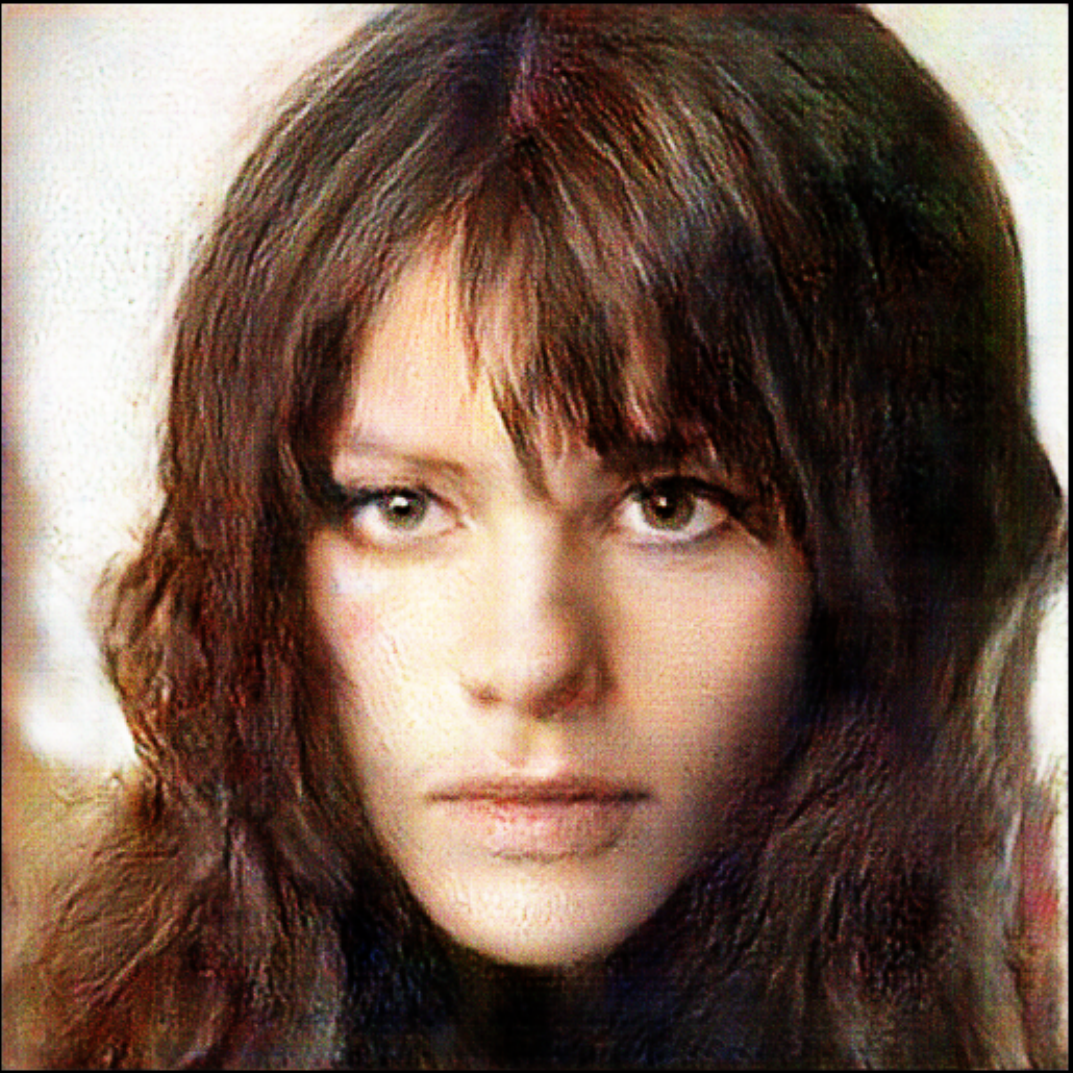}
   \caption{\textbf{HDCGAN Example Result. CelebA-hq}. 229 epochs of training. Image size 512$\times$512. 27\% of full-scale image.}
\label{fgr:hdcgan229}
\end{figure}

\cite{Karras18} introduces CelebA-hq, a set of 30.000 high-definition images to improve training on CelebA. A set of samples generated by HDCGAN on CelebA-hq can be seen in Figures \ref{fgr:celeba_hq}, \ref{fgr:celeba_hq2} and \ref{fgr:hdcgan229}. 

\begin{figure}[H]
\centering
   \includegraphics[scale=0.39]{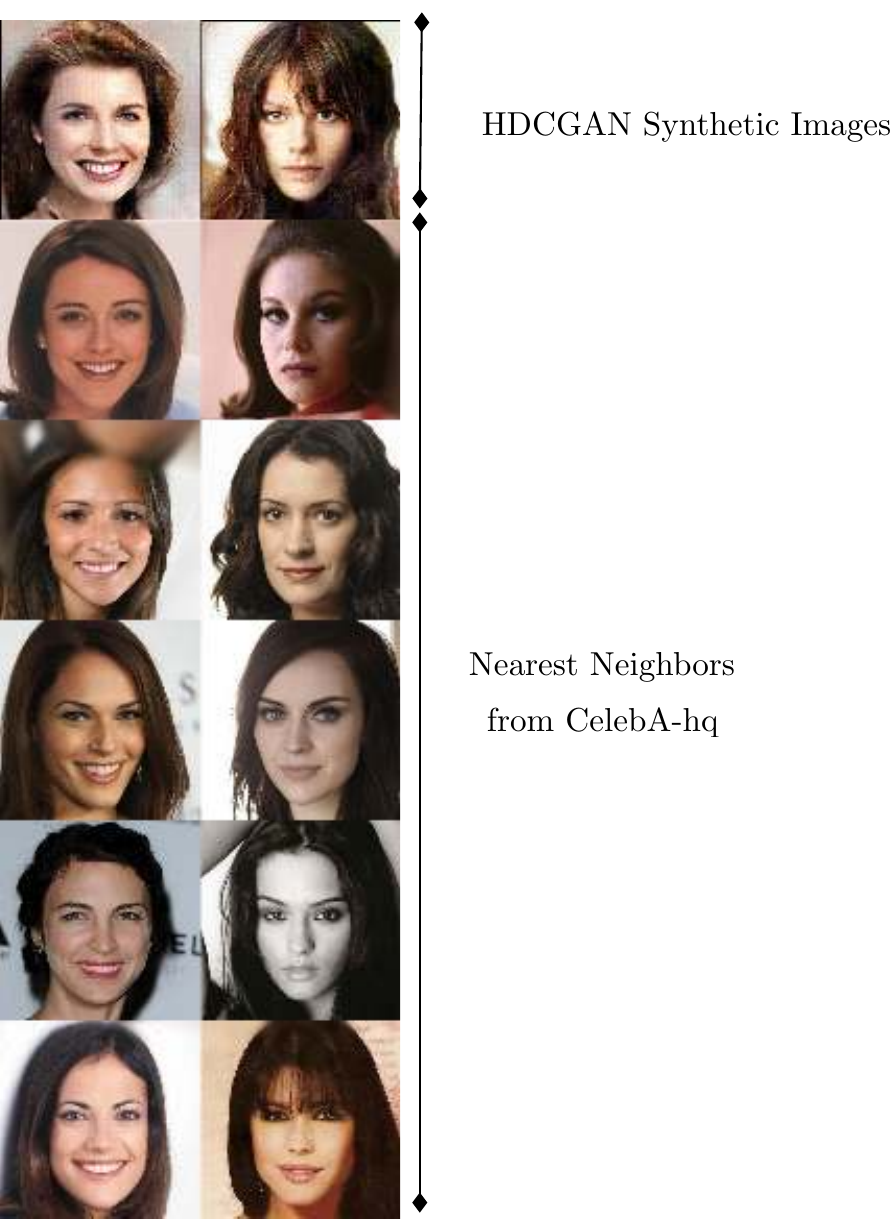}
   \caption{\textbf{Nearest Neighbors. CelebA-hq}. Generated samples in the first row and their five nearest neighbors in training (rows 2-6).}
\label{fgr:nn2}
\end{figure}

To exemplify that the model is generating new bona fide instances instead of memorizing samples from the training set, we retrieve nearest neighbors to the generated images in Figure \ref{fgr:nn2}.

\section{Assessing the Discriminability and Quality of Generated Samples}

We build on previous image similarity metrics to qualitatively evaluate generated samples of generative models. The most effective of these is multi-scale structural similarity (MS-SSIM) \cite{Odena17}. We make comparison at resized image size 128$\times$128 on CelebA. MS-SSIM results are averaged from 10,000 pairs of generated samples. Table \ref{tbl:ms-ssim} shows HDCGAN significantly improves state-of-the-art results.

\setlength{\tabcolsep}{4pt}
\begin{table}[H]
\begin{center}
\caption{Multi-scale structural similarity (MS-SSIM) results on CelebA at resized image size 128$\times$128. Lower is better.}
\label{tbl:ms-ssim}
\begin{tabular}{ll}
\hline\noalign{\smallskip}
  & MS-SSIM   \\
\noalign{\smallskip}
\hline
\noalign{\smallskip}
 \cite{Gulrajani17}  & 0.2854  \\
 \cite{Karras18}  & 0.2838 \\
 HDCGAN & \textbf{0.1978} \\
\hline
\end{tabular}
\end{center}
\end{table}
\setlength{\tabcolsep}{1.4pt}

\begin{figure}[H]
\centering
\includegraphics[scale=0.65]{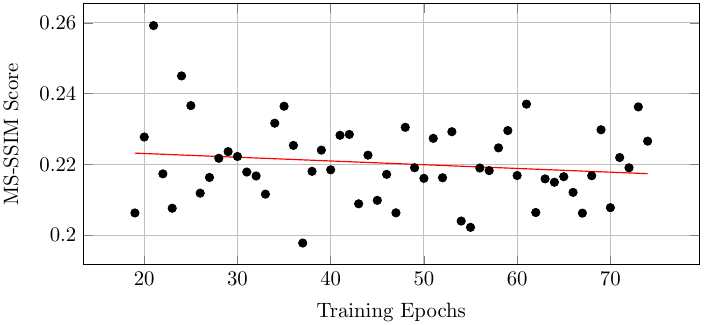}
\caption{\textbf{MS-SSIM Scores on CelebA across several epochs}. Results are averaged from 10,000 pairs of generated images from epoch 19 to 74. Comparison is made at resized image size 128$\times$128. Affine interpolation is shown in red. }
\label{fgr:msssim}
\end{figure}

We monitor MS-SSIM scores across several epochs averaging from 10,000 pairs of generated images to see the temporal performance, Figure \ref{fgr:msssim}. HDCGAN improves the quality of the samples while increases the diversity of the generated distribution.

In \cite{Heusel17} they propose to evaluate GANs using the FR\'ECHET Inception Distance, which assesses the similarity between two distributions by the difference of two Gaussians. We make comparison at resized image size 64$\times$64 on CelebA. Results are computed from 10,000 512$\times$512 generated samples from epochs 36 to 52, resized at image size 64$\times$64 yielding a value of 8.44, Table \ref{tbl:frechet}, clearly outperforming current reported scores in DCGAN architectures \cite{Wu18}.

\setlength{\tabcolsep}{4pt}
\begin{table}[H]
\begin{center}
\caption{FR\'ECHET Inception Distance on CelebA at resized image size 64$\times$64. Lower is better.}
\label{tbl:frechet}
\begin{tabular}{ll}
\hline\noalign{\smallskip}
  & Fr\'echet   \\
\noalign{\smallskip}
\hline
\noalign{\smallskip}
 \cite{Karras18}  & 16.3  \\
 \cite{Wu18}  & 16.0 \\
 HDCGAN & \textbf{8.44} \\
\hline
\end{tabular}
\end{center}
\end{table}
\setlength{\tabcolsep}{1.4pt}


\section{Discussion}

In this paper, we propose High-resolution Deep Convolutional Generative Adversarial Networks (HDCGAN) by stacking SELU $+$ BatchNorm (BS) layers. The proposed method generates high-resolution images (\eg 512$\times$512) in circumstances where the majority of former methods fail. It exhibits a steady and smooth mechanism of training. It also introduces Glasses, the notion that enlarging the input image by a telescope $\zeta$ while keeping all convolutional filters unchanged, can arbitrarily improve the final generated results. HDCGAN is the current state-of-the-art in synthetic image generation on CelebA (MS-SSIM 0.1978 and FR\'ECHET Inception Distance 8.44).
\\

Further, we present a bias-free dataset of faces containing well-balanced ethnical groups, Curt\'o \& Zarza, that poses a very difficult challenge and is rich on learning attributes to sample from. Moreover, we enhance Curt\'o with 4,239 unlabeled synthetic images generated by HDCGAN, being therefore the first GAN augmented dataset of faces.


\bibliographystyle{ACM-Reference-Format}

\end{document}